\documentclass[runningheads]{llncs}
\usepackage{amsmath,graphicx}
\usepackage{amsfonts}
\usepackage{amssymb}
\usepackage{multirow}
\usepackage[table]{xcolor}
\usepackage{hyperref}
\usepackage{comment}
\usepackage{diagbox}
\usepackage{algorithm}
\usepackage{algpseudocode}
\usepackage{xcolor}
\usepackage{color,soul}
\definecolor{light-gray}{gray}{0.85}
\definecolor{orange}{HTML}{E8601C}
\definecolor{green}{HTML}{4EB265}
\definecolor{blue}{HTML}{1965B0}
\usepackage{wrapfig}
\usepackage{subcaption}

\title{Balancing Label Imbalance in Federated Environments Using Only Mixup and Artificially-Labeled Noise}
\usepackage{tikz}
\usetikzlibrary{external}
\begin{document}
%
\author{Tahseen Rabbani* \and
Kyle Sang* \and
Furong Huang}
\authorrunning{Tahseen Rabbani, Kyle Sang, et al.}
\institute{University of Maryland, College Park MD 20742, USA}
\maketitle
\begin{abstract}
Clients in a distributed or federated environment will often hold data skewed towards differing subsets of labels. This scenario, referred to as heterogeneous or non-iid federated learning, has been shown to significantly hinder model training and performance. In this work, we explore the limits of a simple yet effective augmentation strategy for balancing skewed label distributions: filling in underrepresented samples of a particular label class using pseudo-images. While existing algorithms exclusively train on pseudo-images such as mixups of local training data, our augmented client datasets consist of both real and pseudo-images. In further contrast to other literature, we (1) use a DP-Instahide variant to reduce the decodability of our image encodings and (2) as a twist, supplement local data using artificially labeled, training-free 'natural noise' generated by an untrained StyleGAN. These noisy images mimic the power spectra patterns present in natural scenes which, together with mixup images, help homogenize label distribution among clients. We demonstrate that small amounts of augmentation via mixups and natural noise markedly improve label-skewed CIFAR-10 and MNIST training.
\end{abstract}
\textit{To appear in the Proceedings of the $4^{\textrm{th}}$ International Conference 4th International Conference on Pattern Recognition and Artificial Intelligence (ICPRAI 2024). }
\section{Introduction}
\label{introduction}

The federated learning (FL) paradigm has drawn great interest in recent years within the distributed ML community due to its inherent privacy, scalability, and performance across myriad vision and language tasks \cite{li2021survey}. 
A key feature of FL is that clients train their models over privately-held data sans exposure to each other or the central host. While well-known FL algorithms, such as FedAvg \cite{mcmahan2017communication}, perform well when locally-held client datasets are drawn from iid distributions, it is widely known that non-iid client data hinders performance and training. "Non-iidness" is a blanket term for many types of heterogeneity among clients \cite{kairouz2021advances}, including: quantity skew (differing sizes of local datasets) and label-skew (biased label representations), the latter of which is our primary interest.

Label-skew refers to the scenario in which client data is biased towards a size $C$ subset of the total $N$-label set. Federated agents may hold an abundance of examples for one particular class while lacking data for another. While heterogeneous label distribution is a realistic assumption and does not necessarily prevent model convergence \cite{li2019convergence}, the effect on FedAvg-trained model accuracy is quite destructive. Over classical supervised vision tasks such as MNIST and CIFAR-10 classification, heterogeneous label skew towards $C=1,2,$ or $3$ labels can drop test accuracy by $\sim 60\%$ when compared to the iid regime \cite{li2022federated}. 

Inspired by the success of mixup approaches along with recent literature demonstrating vision models can learn by training on synthetic, 'naturalistic' data \cite{baradad2021learning}, our work makes the following contributions: \textbf{\textit{(1)}} We propose local training data-augmentation in a federated environment via our novel DP-LabelHide-mixed \cite{borgnia2021dp} images. In contrast to existing literature, we avoid mixing labels to obfuscate the mixture coefficients of other (non-dominant) labels. \textbf{\textit{(2)}} In addition to mixup images, clients augment underrepresented classes using artificially-labeled, untrained StyleGANv2 images, which we refer to as ``natural noise." \textit{Against initial intuition, these images are not related to their given label.} Rather, the natural power spectra patterns present in the image help to offset problems related to non-iid data. This reimagines the approach of \cite{baradad2021learning} as a private augmentation strategy for federated ecosystems. We validate our approach on CIFAR-10 and MNIST classification over a system of 10 label-skewed clients.

\section{Related Work}
\label{related} 

\textbf{Heterogeneous FL.} FedAvg \cite{mcmahan2017communication}, the first prominent federated algorithm, was empirically shown to converge over non-iid partitions of MNIST without a formal guarantee. Convergence of non-iid training was shown in \cite{li2019convergence} with strong-convex assumptions of each client's local loss function. However, the convergence of non-iid, non-convex training was proven prior to the precise formulation of FL for decentralized learning \cite{lian2017can,lian2015asynchronous}. 

Despite convergence guarantees, the empirical performance of non-iid training is far inferior to iid training. As shown in \cite{li2022federated}, the performance of a CIFAR-10 classifier plummets from 68\%, using a balanced Dirichlet assignment of examples to clients, to 58\%, 50\%, and 10\% for $C=3,2,1$ label-skewed distributions, respectively. The single label skew is especially harmful, regardless of the dataset -- we suspect most tolerable non-iid assumptions, including BCVG, are violated within such non-iid regimes.

\textbf{Data-Augmented FL.} One of the earlier known works to use dataset augmentation to relieve heterogeneity is FedShare \cite{zhao2018federated}, in which the server transmits real examples from a public dataset to clients. While only small number of examples are needed to significantly boost model performance, it is unrealistic to assume the server has examples for every or any label. In lieu of real examples, sample mixing is a popular augmentation strategy. Mixup \cite{zhang2017mixup} forms convex combinations of $k$ examples ($k$-way mixup) and their labels, with weights drawn from common random distributions. Mixup and its privacy-improved variant InstaHide \cite{huang2020instahide} produces pseudo-synthetic training data beneficial to improving model-robustness against adversarial examples. 

FedMix \cite{yoon2021fedmix} was one of the earlier known works to use mixup to balance label-skew. To homogenize data distributions, $k$-way mixups of all client data and labels are constructed and sent up to the server, where it is then unioned and sent back down to clients for training. This approach can prove expensive over large client clusters, especially if the total data size across all clients is massive, and for smaller, $C=1$ label-skewed clients, their mixups are easily decodable by other clients with rich, balanced data \cite{carlini2021private} due to information gleaned from the "pseudo"-label -- which will contain one entry. XORMixFL \cite{shin2020xor} also distributes mixed-client data via mixup and is intended for one-shot server-side training, as opposed to directly improving local training.

\section{Preliminaries}
\label{prelim}

\subsection{DP-InstaHide}
\begin{figure*}
  \centering

  \begin{subfigure}[t]{0.65\textwidth} 
    \centering
    \includegraphics[width=\textwidth]{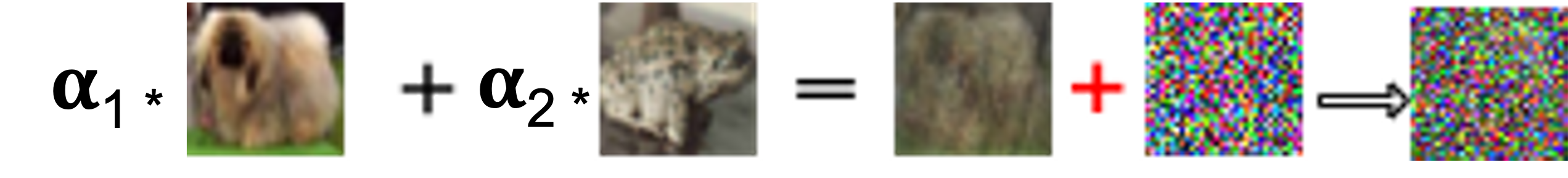}
    \caption{DP-LabelHide}
    \label{fig:cifarmix}
  \end{subfigure}
  \hspace{1cm} 
  \begin{subfigure}[t]{0.25\textwidth} 
    \centering
    \includegraphics[width=\textwidth]{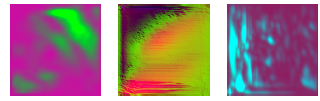}
    \caption{Natural Noise}
    \label{fig:natural-noise}
  \end{subfigure}
  \caption{\textbf{Pseudo-images.} We augment local training data with two varieties of pseudo-images: mixup and natural noise. Figure \ref{fig:cifarmix} depicts 2-way mixup of CIFAR-10 using Algorithm \ref{alg:imagemixup}. Figure \ref{fig:natural-noise} depicts unlabeled, StyleGAN-Oriented natural images.}
  \label{fig:mixing}
\end{figure*}

\begin{algorithm}
\caption{DP-LabelHide}
\label{alg:imagemixup}
\begin{algorithmic}[1]
\Require Training dataset $\mathcal{D} \subseteq X\times Y$ (where $X\in\mathbb{R}^d)$, mixup parameter $k$, target label $j$, Laplacian parameter $\sigma$
\State Randomly select an image $(x_1,j)\in\mathcal{D}$ with label $j$.
\State Randomly select $k-1$ other images $\{(x_i,y_i)\}_{i=2}^{k}\subset \mathcal{D}$ without replacement.
\State Randomly draw a $k$-dimensional unit vector $\alpha$. Sort entries of $\alpha$ in a decreasing manner.
\State Generate mixup: $x_{\text{mixed}} \leftarrow \sum_{i}^k \alpha_i x_i$
\State Draw $\eta\sim Lap(\textbf{0},\sigma I)\in \mathbb{R}^d$.
\State DP mixup: $x_{\text{DPMix}} \leftarrow x_{\text{mixed}}+\eta$
\Return{$(x_{\text{DPMix}},j)$}
\end{algorithmic}
\end{algorithm}
Our clients will distribute $k$-way mixes of their local samples to one another using our own variation of DP-InstaHide which we refer to as DP-LabelHide, described in Algorithm \ref{alg:imagemixup} and visualized in Figure \ref{fig:mixing}.  There are a few notable differences between Algorithm \ref{alg:imagemixup} and the conventional DP-InstaHide. Our algorithm is designed to serve a mixed sample encoding a target label, as opposed to purely random mixes. As opposed to taking a pure average of $k$ samples, we form a stochastic combination, with the heaviest weight going to the sample belonging to the target class. We do so to maximally encode features strongly correlated with label $j$ in the mixup. Additionally, we do not mix the label; we simply fix the label as the target label. We do so to hide the presence of other labels available within the mixer's dataset.

A key feature of DP-Insta/LabelHide is the addition of an independent $d$-dimensional isotropic Laplacian noise vector. The corresponding density function is $f_{\sigma}(\eta)=\frac{1}{(2\sigma)^d}e^{||\eta||_1/\sigma}$. The randomized mechanism $\mathcal{M}$ which produces mixed images via DP-InstaHide (and by corollary, our DP-LabelHide) is differentially-private and provides a strong defense against data poisoning; see \cite{borgnia2021dp}. 

\subsection{StyleGAN}
We will also augment client data using generative adversarial networks (GANs). Following the approach of \cite{baradad2021learning}, we will use an untrained StyleGANv2 \cite{karras2020analyzing} to produce data encoding natural-image-like properties. Randomly initialized deep architectures, including GANs and deep CNNs, encode statistical priors inherent to natural images. Such priors can be used to generate synthetic data \cite{ruderman1994statistics} which can be used to perform low-level vision tasks \cite{ulyanov2018deep}. 

Different initializations of noise maps (hallmark structures of GANs), convolutional filters, and biases applied to output channels can produce natural images with varied patterns. For example, when supplied with random, normalized inputs, differing initializations can coerce output images to contain features such as high frequency, sparsity, and oriented structures. StyleGAN-Oriented \cite{baradad2021learning} (which we use in our experiments) uses the following initialization for an untrained StyleGANv2 architecture \cite{karras2020analyzing}: randomly sample a wavelet filter $f_i$. The convolution is then set as $y_k=\sum_{i}[\alpha_{k,i}(x_i \star f_i)]+\beta_k$, where $k$ denotes the output channel, $i$ is the input channel,  $\alpha_{k,i}\sim \mathcal{N}(0,1)$ is an amplitude multiplier, and $\beta_k\sim \mathcal{U}(-0.2, 0.2)$ is a bias term. Note that $f_i=f$ for all $i$; the wavelet is shared across all output channels. Clients will label these training-free natural output images with deficient labels.

\section{Augmentation via DP-LabelHide and Natural Noise}
\label{balance}
In this section, we detail our protocol for balancing label-skewed federated environments. 
\begin{algorithm}
\caption{Label Balance}
\label{alg:label-balance}
\begin{algorithmic}[1]
\Require Local data $\mathcal{D}$, target label $j$, Number of required examples $P$, mixup proportion $\lambda$
\State Received examples $E\leftarrow 0$
\State Client issues request for $\alpha\cdot P$ examples with label $j$
\State Client receives a set $\mathcal{D}_{\mathrm{mixed}}$ of examples with label $j$
\If{$|S|>\alpha\cdot P$}
    \State Client randomly trims $S$ down to size $\alpha\cdot P$
\EndIf
\State $E\leftarrow \min(|S|,\alpha\cdot P)$
\State Client generates a set $\mathcal{D}_{\mathrm{noise}}$ of $P-E$ natural noise images labeled {$j$}.
\State $\mathcal{D}\leftarrow \mathcal{D}\cup \mathcal{D}_{\mathrm{mixed}}\cup \mathcal{D}_{\mathrm{noise}}$
\Return{$\mathcal{D}$}
\end{algorithmic}
\end{algorithm}

A client, the requester, first determines a target underrepresented label $j$ and how many examples $P$ they need. While $P$ can reflect the size of their maximally represented class, we demonstrate in Section \ref{experiments} that far fewer examples are needed. The client also determines what proportion $\alpha$ of $P$ should consist of mixups. Once this is decided, the client issues a public bounty for $\alpha\cdot P$ samples with label $j$. The client may dispatch this request through the central server or a trusted and connected peer in a pre-defined communication topology as in decentralized FL \cite{kairouz2021advances}. 

Clients on the receiving end of the request determine how many mixups with label $j$ they are willing to distribute to the requester. This will depend on two key factors, including: the local availability of examples with label $j$ and the presence of enough samples living in other classes to create a $k$-way mixup. If the receiver is willing, they proceed to create mixed examples via Algorithm \ref{alg:imagemixup}. The mixed samples are either distributed directly to the requester (if they have a trusted communication edge) or to the server, who then routes the mixings to the requester. 

After an internally-defined wait period, the requester aggregates all transmitted mixed samples. Any remaining proportion of $P$ is filled by natural noise images labeled as class $j$. The client may use any training-free generative strategy they so desire, such as StyleGAN-Oriented initialization, or receive unlabeled, natural noise images from the server. This process may be repeated for other underrepresented classes until there is sufficient representation across all labels.

\textbf{Remarks.}\textit{(1)} In synchronized FL, this label request represents a blocking communication. Clients should opt to choose shorter waiting periods, especially since they can generate all $P$ examples themselves via natural noise, albeit with limited improvements. We observe in Section \ref{experiments} that no amount of natural noise augmentation will harm base-level non-iid performance.
\textit{(2)} The quality of mixups decreases with the number of ingredient images $k$ in the mix \cite{yoon2021fedmix,borgnia2021dp,zhang2017mixup}.
\textit{(3)} The split between mixed images and natural noise is a client decision, although our experiments suggest that $50/50$ or $75/25$ splits between mixed/natural images are acceptable and sufficient.
\textit{(4)} A popular mixup approach, FedMix \cite{yoon2021fedmix}, replace all local training data with an amalgamation of mixups across all client images. Our approach is amenable to partial participation and does not require any training data replacement.

\section{Experiments}
\label{experiments}

In this section, we examine the effectiveness of our federated augmentation strategy in training CIFAR-10 and MNIST classifiers. \textit{We observe that natural noise and mixup are synergistic augmentation strategies and we do not necessarily require large amounts of these pseudo-images to boost performance.}

\textbf{Experimental Setup.} For both the CIFAR-10 and MNIST classification tasks we train a ResNet-18 architecture \cite{he2016deep}. We use 10 clients who all train for 1 epoch (batch size 128) before models average according to the FedAvg \cite{mcmahan2017communication} protocol. An Adam optimizer with a learning rate of 0.001, $\beta_1=0.9$, and $\beta_2=0.999$ is used via PyTorch on a high-performance computing cluster employing A4000 GPUs. Prior to augmentation, each client only has access to disparate training data corresponding to one, two, or three classes ($C=1$, $C=2$, $C=3$). 

\textbf{Augmentation Setup.} To supplement images in deficient classes we simulate Algorithm \ref{balance} using combinations of mixup images and natural noise. 

\textit{Mixup images:} If client $i$ is missing an example for label $j$, we employ a 4-way mixup according to Algorithm \ref{alg:imagemixup}, by combining a class $j$ training image and 3 other random images. We set the dominant weight $\alpha_1 \sim \mathcal{U}[0.5,0.75]$. The remaining weights are randomly selected to sum to $1-\alpha_1$, ensuring the target image encodes 50\% to 75\% of the mixup. We perform the weighted sum over the images and then add a noise vector $\eta \sim Lap(0,50I)$. 

\textit{Natural noise:} We use an untrained StyleGANv2 model to produce oriented natural images due to research indicating this is the most successful style of natural noise for learning low-level vision. See Figure \ref{fig:natural-noise} for examples. We use publicly-released code by the authors of \cite{baradad2021learning} to initialize StyleGAN-Oriented: \url{https://github.com/mbaradad/learning_with_noise}. 

\subsection{MNIST Augmentation}
\begin{wraptable}{r}{0.55\linewidth}
\begin{tabular}{|l|l|}
\hline
\textbf{MNIST}             & \textbf{10\% Supplement} \\ \hline
100\% Mixup/ 0\% Natural & 96.96202532              \\ \hline
75\% Mixup/ 25\% Natural & 93.67088608              \\ \hline
50\% Mixup/ 50\% Natural & 93.92405063              \\ \hline
25\% Mixup/ 75\% Natural & 94.05063291              \\ \hline
0\% Mixup/ 100\% Natural & 60.25316456              \\ \hline
No Supplement         & 23.5443038               \\ \hline
IID                  & 98.60759494              \\ \hline
\end{tabular}
\caption{\textbf{MNIST Augmentation.} Every augmentation strategy significantly boosts model performance and natural noise alone increases performance by nearly 40\%.}
\label{tab:mnist-top}
\end{wraptable}
For MNIST experiments, we used 10\% supplementation (600 mix/nat images per missing class) on 200 communication rounds. Our non-iid baseline hovers around 23\%. According to Table \ref{tab:mnist-top}, models trained with supplements have significantly boosted performance over the non-iid baseline. \textit{Even if the supplement consists only of natural noise images performance is boosted by nearly 40\%, indicating that artificially-labeled noise provides training benefits in label-skewed FL.} Usage of mixup increases performance to near-IID performance. 
 


Although the physical presentation of the oriented natural images somewhat resembles MNIST (clustering of bright pixels against a darker background), they are indiscriminately labeled.  Hence, the resemblance between the feature distributions alone cannot account for the drastic boost in performance we observe in Table \ref{tab:mnist-top} when only using natural noise.

\begin{table*}
\centering
\begin{tabular}{|l|c|c|c|c|c|}
\hline
Mix/Nat Split & 100\% Supplement & 20\% Supplement & 10\% Supplement \\
\hline
100\% Mix/0\% Nat & \textcolor{orange}{37.03}/\textcolor{green}{62.22}/\textcolor{blue}{71.39} & \textcolor{orange}{42.66}/\textcolor{green}{62.34}/\textcolor{blue}{72.03} & \textcolor{orange}{33.48}/\textcolor{green}{61.39}/\textcolor{blue}{73.42}\\
\hline
75\% Mix/25\% Nat & \textcolor{orange}{44.75}/\textcolor{green}{61.58}/\textcolor{blue}{72.53} & \textcolor{orange}{\textbf{\hl{45.32}}}/\textcolor{green}{64.05}/\textcolor{blue}{74.81} & \textcolor{orange}{34.56}/\textcolor{green}{\textbf{\hl{66.46}}}/\textcolor{blue}{77.15} \\
\hline
75\% Mix/0\% Nat & \textcolor{orange}{39.75}/\textcolor{green}{62.15}/\textcolor{blue}{69.49} & \textcolor{orange}{33.16}/\textcolor{green}{61.65}/\textcolor{blue}{73.54} & \textcolor{orange}{26.33}/\textcolor{green}{51.77}/\textcolor{blue}{77.02} \\
\hline
50\% Mix/50\% Nat & \textcolor{orange}{42.09}/\textcolor{green}{\textbf{\hl{65.00}}}/\textcolor{blue}{74.62} & \textcolor{orange}{41.71}/\textcolor{green}{\textbf{\hl{64.94}}}/\textcolor{blue}{73.42} & \textcolor{orange}{\textbf{\hl{38.92}}}/\textcolor{green}{64.94}/\textcolor{blue}{74.49}\\
\hline
50\% Mix/0\% Nat & \textcolor{orange}{\textbf{\hl{49.05}}}/\textcolor{green}{64.49}/\textcolor{blue}{71.65} & \textcolor{orange}{36.84}/\textcolor{green}{59.62}/\textcolor{blue}{68.99} & \textcolor{orange}{20.13}/\textcolor{green}{57.34}/\textcolor{blue}{74.43}\\
\hline
25\% Mix/75\% Nat & \textcolor{orange}{31.52}/\textcolor{green}{55.95}/\textcolor{blue}{\textbf{\hl{74.94}}} & \textcolor{orange}{41.58}/\textcolor{green}{63.54}/\textcolor{blue}{\textbf{\hl{77.97}}} & \textcolor{orange}{29.11}/\textcolor{green}{63.54}/\textcolor{blue}{\textbf{\hl{77.97}}}\\
\hline
25\% Mix/0\% Nat & \textcolor{orange}{48.10}/\textcolor{green}{59.24}/\textcolor{blue}{73.92} & \textcolor{orange}{19.75}/\textcolor{green}{63.54}/\textcolor{blue}{75.57} & \textcolor{orange}{13.42}/\textcolor{green}{62.28}/\textcolor{blue}{76.58}\\
\hline
0\% Mix/100\% Nat & \textcolor{orange}{17.85}/\textcolor{green}{46.90}/\textcolor{blue}{46.90} & \textcolor{orange}{15.89}/\textcolor{green}{58.99}/\textcolor{blue}{74.11} & \textcolor{orange}{17.72}/\textcolor{green}{63.54}/\textcolor{blue}{77.59}\\
\hline
\cellcolor{light-gray}\textbf{No Supplement} & \multicolumn{3}{c|}{\cellcolor{light-gray}\textcolor{orange}{14.11}/\textcolor{green}{52.53}/\textcolor{blue}{69.62}}\\
\hline
\cellcolor{light-gray}\textbf{IID} & \multicolumn{3}{c|}{\cellcolor{light-gray}87.97} \\
\hline
\end{tabular}
\caption{\textbf{CIFAR-10 Augmentation.} Ablation results for \textcolor{orange}{ $C=1$}/\textcolor{green}{ $C=2$}/\textcolor{blue}{ $C=3$} class skew across different variations of supplement size and mixup/natural image splits. For example, a 100\% supplementation on $C=1$ would supplement 5000 pseudo-images to a client's 9 missing classes. Hybrid augmentations of mixup and natural noise generally perform the best. }
\label{cifar10full}
\end{table*}

\subsection{CIFAR-10 Augmentation}
For CIFAR-10 experiments, our ablations had three dimensions: the size of the supplement, the split of natural and mixed images, and class skew ($C=1,2,3$) per client. Results are depicted in Table \ref{cifar10full}.

\begin{figure}
  \vspace{-1em}
  \centering

  \begin{subfigure}[b]{0.41\textwidth}
    \includegraphics[width=\textwidth]{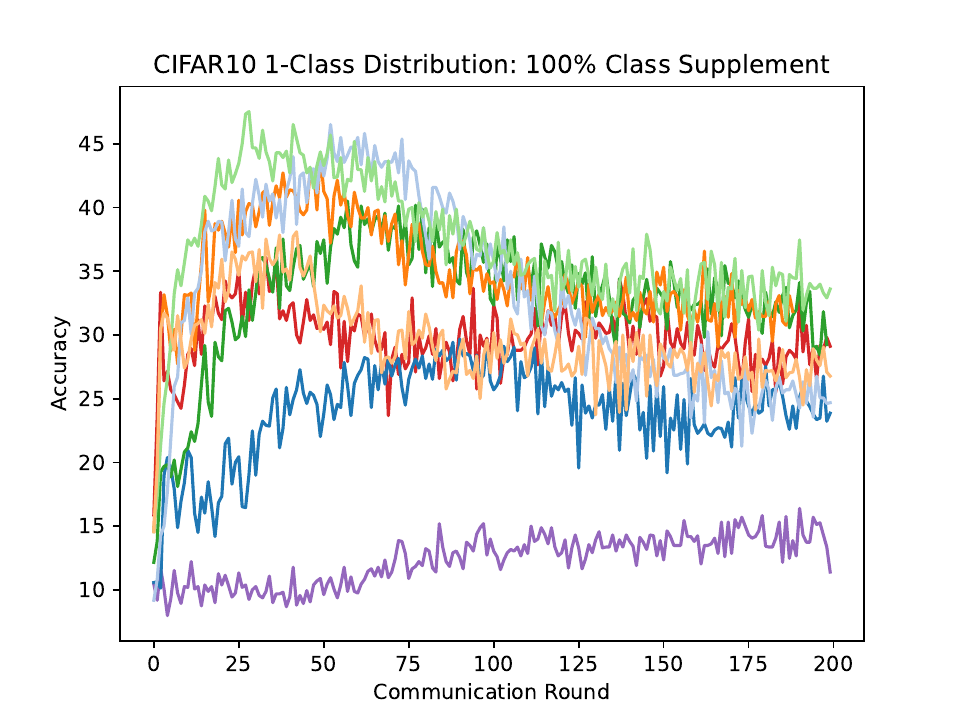}
    \caption{100\% Supplement}
    \label{fig:exclusion-100}
  \end{subfigure}
  \begin{subfigure}[b]{0.41\textwidth}
    \includegraphics[width=\textwidth]{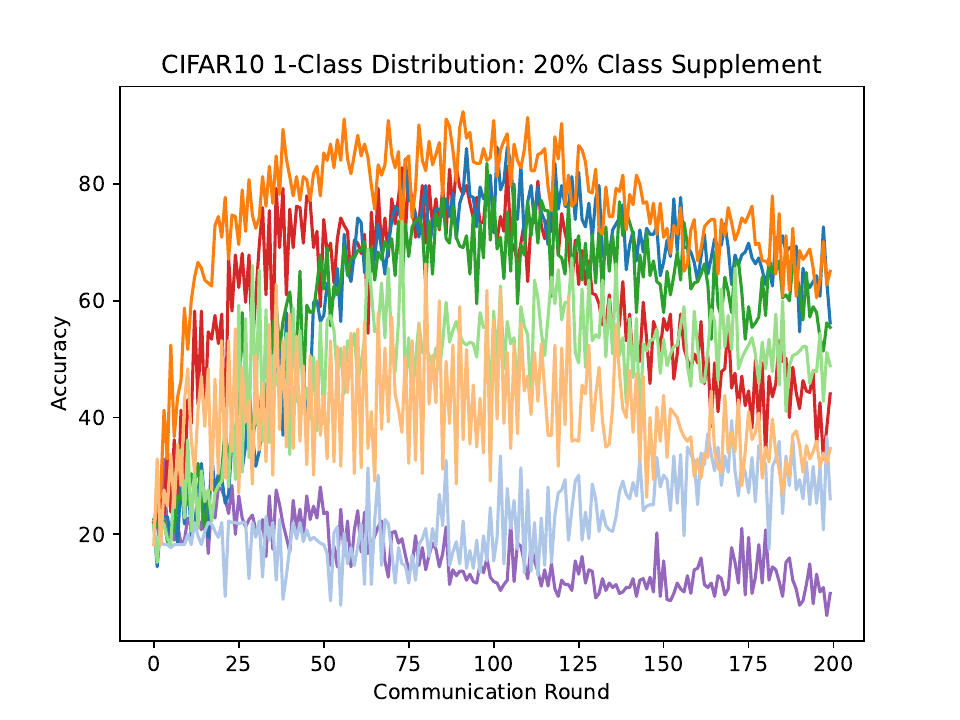}
    \caption{20\% Supplement}
    \label{fig:exclusion-20}
  \end{subfigure}
  \begin{subfigure}[b]{0.41\textwidth}
    \includegraphics[width=\textwidth]{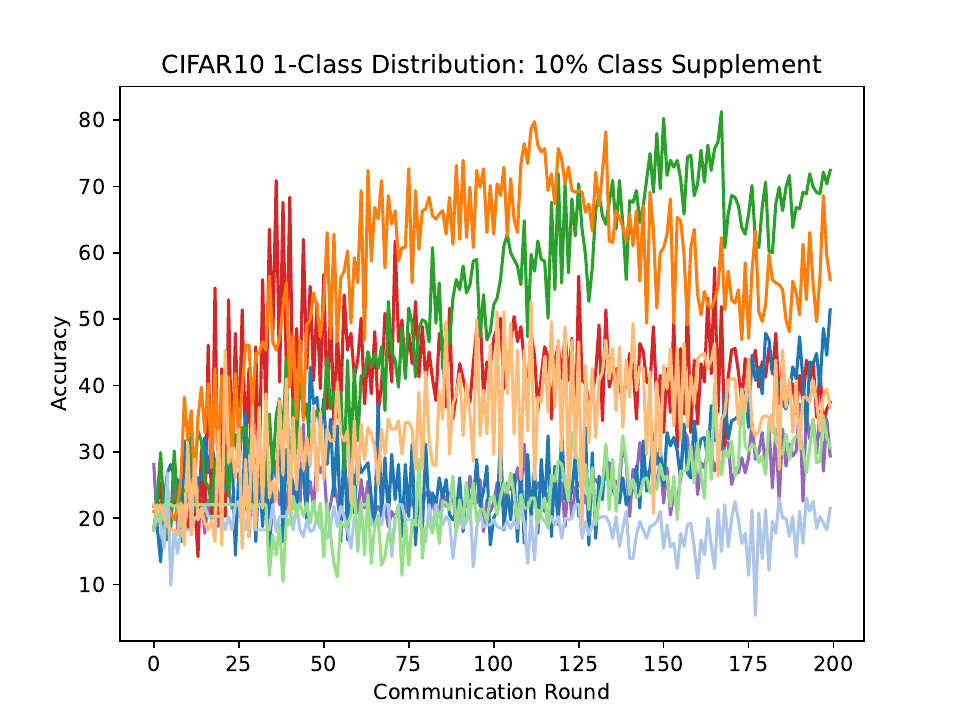}
    \caption{10\% Supplement}
    \label{fig:exclusion-10}
  \end{subfigure}
  \begin{subfigure}[b]{0.41\textwidth}
    \includegraphics[width=\textwidth]{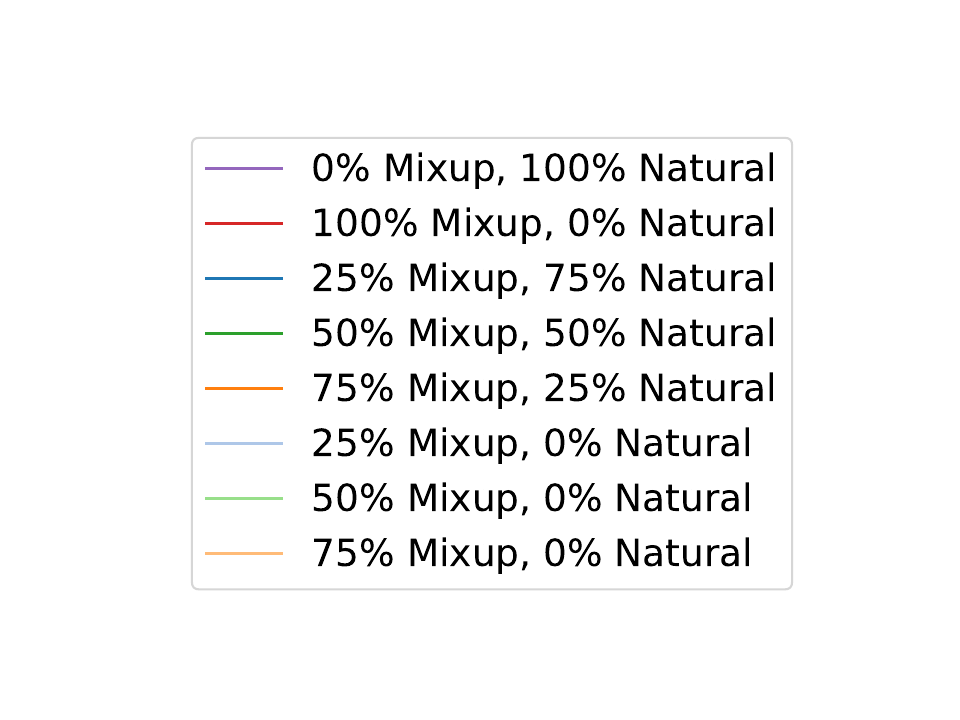}
    \label{fig:legend}
  \end{subfigure}

    \caption{\textbf{Extreme Label-Imbalance.} A visualization of $C=1$ label skew. Hybrid strategies (mixup + natural noise) generally outperform supplements consisting solely of mixups. Natural images boost performance at lower supplements. 
    }
  \label{fig:mainfig}
\end{figure}

The baseline (no supplement, $C=1$) for CIFAR-10 hovers around 10\%. Using a hybrid augmentation natural images and mixed-up images almost unanimously yields better results, with 75\% and 50\% mixup often outperforming 100\% mixup or 0\% mixup (see Figure \ref{fig:mainfig}).  \textit{Excluding natural images has deleterious effects at lower supplementation amounts. In tandem, mixup and natural noise are complementary augmentations.}

As the number of classes per client increases ($C=2, C=3$), performance boosts from supplementation seem to lessen as the overall baseline test accuracy significantly rises. 100\%, 20\%, and 10\% supplementation seem to converge towards similar model accuracies, with diminishing returns as supplementation increases. As $C$ increases, the "required" amount of supplementation for optimal performance seems to decrease, as well as favoring higher proportions of natural images as shown by $C=3$. Additionally, it is not necessary to use supplements of tremendous size to observe significantly boosted model performance. \textit{Augmented sets which are 10\% the size of available real data is enough to reach high levels of performance.}

Across all our ablations, we observe these same trends of low supplementation requirements, especially when there are fewer missing classes. The vital balance of natural images and mixup images also suggests the natural power spectra present in natural images play a key role in achieving better performance. We theorize that these natural spectra images homogenize loss landscapes, preventing the model from overfitting to the extremely noisy mixed images. When observing other tasks and similar baselines, this trend continues as seen in \ref{baselines}, as it performs best when there are equal or more natural images in proportion to mixed images. While the proportion for every task is different, it seems that less complex tasks require less supplementation and less skewed data benefits from a higher proportion of natural images.

\vspace{-2em}
\begin{table}[htbp!]
  \centering
  \caption{Comparison of Augmentation Methods}
    \begin{tabular}{|c|c|c|c|c|c|c|}
    \hline
    \multirow{2}[4]{*}{Method} & \multicolumn{2}{c|}{CIFAR-10} & \multicolumn{2}{c|}{CIFAR-100} & \multicolumn{2}{c|}{SVHN} \\
\cline{2-7}          & $\alpha=0.05$ & $\alpha=0.1$ & $\alpha=0.05$ & $\alpha=0.1$ & $\alpha=0.05$ & $\alpha=0.1$ \\
    \hline
    \hline
    75\% mixup 25\% natural & 64.28 & {\textbf{\hl{69.34}}} & 40.9  & 42.98 & 93.24 & 93.88 \\
    50\% mixup 50\% natural & {\textbf{\hl{65.27}}} & 68.11 & 42.8 & 45.03 & 94.56 & 94.67 \\
    25\% mixup 75\% natural & 63.11 & 68.51 & {\textbf{\hl{43.47}}} & {\textbf{\hl{45.85}}} & {\textbf{\hl{94.99}}} & {\textbf{\hl{94.76}}} \\
    FedLC                   & 54.55 & 65.91 & 38.08 & 41.01 & 82.36 & 84.41 \\
    FedRS                   & 44.39 & 54.04 & 27.93 & 32.89 & 75.97 & 83.27 \\
    \hline
    \end{tabular}%
  \caption{\textbf{Competitor Baselines.} We compare our method (20\% supplementation) against two popular competitors, FedLC \cite{zhang2022federated} and FedRS \cite{li2021fedrs} with the same settings (non-iid Dirichlet, 20 clients, 400 communication rounds). FedLC and FedRS numbers are reported from \cite{zhang2022federated}. We find across several tasks, our method performs better. }
\label{baselines}
\end{table}%

\section{Conclusion}
\label{conclusion}
In this work, we alleviate extreme label-skew in a federated environment by supplementing local client data with pseudo-images consisting of mixups and artificially-labeled natural noise. These augmentations are included alongside real examples within the training data. Either variety of augmentation significantly boosts global model performance on CIFAR-10, CIFAR-100, SVHN, and MNIST classification tasks, and we observe synergy between both strategies. Our framework provides an efficient and private strategy for balancing extreme label-imbalance in a distributed environment. 

\bibliographystyle{splncs04}
\bibliography{supp_bib}

\end{document}